\documentclass[letterpaper, 10 pt, conference]{ieeeconf}  

\IEEEoverridecommandlockouts                              

\usepackage{booktabs}
\usepackage{graphicx}
\usepackage{url}
\usepackage{subfigure}
\usepackage{amsmath}
\usepackage{amsfonts}
\usepackage{adjustbox}

\usepackage{algorithm}
\usepackage{algpseudocode}
\usepackage{amsmath}

\title{\LARGE \bf
TOCALib: Optimal control library with interpolation for bimanual manipulation and obstacles avoidance
}

\author{Yulia Danik$^{1}$, Dmitry Makarov$^{2}$, Aleksandra Arkhipova$^{3}$,
Sergei Davidenko$^{4}$ and Aleksandr Panov$^{5}$  
\thanks{$^{1}$Yulia Danik is with FRC CSC RAS, Moscow, Russia and MIPT, Dolgoprudny, Russia
        {\tt\small yuliadanik@gmail.com}}%
\thanks{$^{2}$Dmitry Makarov is with 
FRC CSC RAS, Moscow, Russia and MIPT, Dolgoprudny, Russia
        {\tt\small makarov@isa.ru}}%
\thanks{$^{3}$ Aleksandra Arkhipova is with Robotics Center, Sberbank of Russia, Moscow, Russia
        {\tt\small alsearkhipova@sberbank.ru}}
\thanks{$^{4}$ Sergei Davidenko is with the Robotics Center, Sberbank of Russia, PhD student in Skolkovo Institute of Science and Technology, Moscow, Russia
        {\tt\small Sergei.Davidenko@skoltech.ru}}
        \thanks{$^{5}$Aleksandr Panov is with the AIRI and  Moscow Institute of Physics and Technology, Moscow, Russia
        {\tt\small panov@airi.net}}
}

\begin{document}
\maketitle
\thispagestyle{empty}
\pagestyle{empty}

\begin{abstract}

The paper presents a new approach for constructing a library of optimal trajectories for two robotic manipulators, Two-Arm Optimal Control and Avoidance Library (TOCALib)\footnote{\url{https://sites.google.com/view/tocalib?usp=sharing}}. The optimisation takes into account kinodynamic and other constraints within the FROST framework. The novelty of the method lies in the consideration of collisions using the DCOL method, which allows obtaining symbolic expressions for assessing the presence of collisions and using them in gradient-based optimization control methods. The proposed approach allowed the implementation of complex bimanual manipulations. In this paper we used Mobile Aloha as an example of TOCALib application. The approach can be extended to other bimanual robots, as well as to gait control of bipedal robots. It can also be used to construct training data for machine learning tasks for manipulation.

\end{abstract}

\section{INTRODUCTION}
The field of robotics is rapidly evolving, with a growing emphasis on the development and deployment of bimanual robots, such as the TIAGo \cite{2016tiago} and anthropomorphic robots like Baxter \cite{sorell2022cobots}, Figure 01, Astribot S1 etc. These robots, designed with two arms to mimic human coordination, have opened up new possibilities to perform complex tasks that require simultaneous control of multiple manipulators. Bimanual tasks are becoming increasingly important as industries seek more sophisticated automation solutions that can handle tasks requiring precise coordination and flexibility, such as assembly, packaging, and collaborative work with humans. Collision avoidance in robot manipulators is essential for safe and efficient operation. To accomplish this task in real time, it is necessary to plan the collision-free paths and control trajectories of the manipulators from the current positions to the desired positions that are comfortable for safe object grasping.

The recent emergence of relatively low-cost mobile bimanual robots has attracted considerable interest from researchers in this field. One of these robots is a Mobile Aloha, which was developed in 2024. It is a mobile platform with two independently controlled wheels equipped with four manipulators (two rare for teleoperation and two front for manipulation). Its manipulators feature 16 DOF in total, with 8 DOF for each arm \cite{Fu2024}.

Usually Aloha is controlled using behavior cloning based on the Action Chunking with Transformers approach \cite{Fu2024}. However, it is still relevant to control Mobile Aloha using alternative methods capable of solving tasks without human intervention and finding optimal or unconventional solutions. These include "classical" optimal control methods and reinforcement learning (RL) approaches. For the implementation of offline RL training or for the practical application of certain online RL methods (e.g. SERL \cite{SERL}), high-quality task solution examples are required.

Thus, for both classical real-time optimal control manipulation and RL-based manipulation, it is desirable to have good initial approximations for the motion of the manipulators. To achieve this, a precomputed motion library has been proposed. The library contains optimal trajectories and controls for the manipulators, taking into account self-collisions and collisions with static and dynamic objects. Each trajectory is parameterised by the initial and final state of the manipulator. 

This paper proposes a method for constructing a precomputed Two-Arm Optimal Control and Avoidance Library (TOCALib) for the manipulators. We use the Mobile Aloha as an example of TOCALib application. The advantages of the library are following:
\begin{itemize}
\item Optimal trajectories and controls taking into account detailed kinodynamic model of the robot.
\item Solution for arbitrary endpoints through the interpolation mechanism (trilinear interpolation) and the use of library entries. 
\item Reduction of computation and search time for trajectories between library grid nodes (no need to solve the optimization problem for new points).
\end{itemize}

The developed software platform can be used to build training datasets for RL agents, as well as a baseline solution to evaluate their performance.

\section{Related works}

\subsection{Trajectory Planning for Manipulators}
We proposed TOCALib, an original approach based on creating a precomputed motion library for manipulators using the FROST package in MATLAB, solving nonlinear programming (NLP) tasks for various end-effector positions with IPOPT and considering a set of symbolic constraints (dynamics, kinematics, collisions). For collision avoidance, the Differentiable Collisions library in Julia is separately integrated, allowing distance evaluation between primitives and obtaining the first derivative of this distance. TOCALib approach is positioned as a means of obtaining a high-quality dataset with manipulator trajectories. Table \ref{tab:precomputed} presents a comparison of this approach with most popular alternative methods based on different criteria. Thus, the proposed approach provides high computational efficiency and physical accuracy through symbolic optimization and consideration of the full dynamic model. These properties make it especially suitable for complex industrial tasks and the generation of high-quality data for reinforcement learning (RL).

\begin{table*}
\caption{Comparison of approaches to creating a precomputed motion library for manipulators}
\setlength{\abovecaptionskip}{0cm}
\label{tab:precomputed}
\scriptsize
\begin{minipage}{\textwidth}
\begin{tabular}{|p{2.2cm}|p{2.5cm}|p{2.5cm}|p{2.5cm}|p{3.2cm}|p{2.5cm}|}
\toprule
\hline
\textbf{Criterion} & \textbf{TOCALib} & \textbf{MoveIt / Tesseract} & \textbf{RoboDK} & \textbf{RMP~Flow\footnote{https://docs.omniverse.nvidia.com/isaacsim/latest/concepts/motion\_generation/rmpflow.html}/TrajOpt\footnote{https://github.com/tesseract-robotics/trajopt?ysclid=m3hqz7burk86695758}/CHOMP\footnote{https://personalrobotics.cs.washington.edu/publications/zucker2013chomp.pdf}} & \textbf{OMPL} \\
\hline
\textbf{References} & This paper & https://moveit.ai/ https://github.com/tesseract-ocr/tesseract   & https://robodk.com/ &  See below the table & https://ompl.kavrakilab.org/ \\
\hline
\textbf{Approach Characteristics} & Symbolic optimization; IPOPT solver for precise trajectories with dynamics and collision considerations; integration with Julia for distance gradient computation & Kinematic planning (OMPL: RRT, PRM); collision avoidance via TrajOpt and DART; tight integration with ROS & Offline programming for industrial tasks & Numerical optimization (SQP, Adam); FCL and Bullet for reactive collision avoidance; efficient in real-time & Heuristic path construction (PRM, RRT), user-defined constraint support \\
\hline
\textbf{Planning Methods} & Nonlinear programming (NLP) with IPOPT and symbolic derivatives & Sampling-based planning with collision constraints (RRT, PRM) & Calculation of static trajectories without optimization & Numerical trajectory optimization (SQP, Adam) & Sampling-based planning (RRT, PRM) \\
\hline
\textbf{Collision Algorithm} & Differentiable Collisions (Julia), symbolic constraints & FCL (Flexible Collision Library) & Simple collision checking for offline tasks & Bullet / FCL (reactive collision avoidance) & FCL for basic collision checking \\
\hline
\textbf{Dynamic Consideration} & Full (forces, torques, and acceleration constraints) & None (kinematics only) & None (only basic motion constraints) & Yes (acceleration and velocity consideration) & None, focused on static planning \\
\hline
\textbf{Constraint Handling} & Symbolic constraints for collisions and dynamics & Collisions and motion boundaries & Range of motion constraints & Collisions, speeds, and accelerations & Basic collision prevention constraints \\
\hline
\textbf{Trajectory Accuracy} & High, symbolic derivatives and dynamic constraints & Medium, limited by kinematic parameters & High for industrial tasks with fixed trajectories & High, adaptive trajectories for dynamic conditions & Low, primitive heuristics \\
\hline
\textbf{Adaptability to Environmental Changes} & Limited, designed for precomputed trajectories & Limited, suitable for static tasks & Low, designed for predefined motions & High, adaptive real-time response & Low, focused on tasks without dynamics \\
\hline
\textbf{Symbolic Constraint Support} & Yes, supported in Julia, useful for gradient optimization & No, numerical methods only & No, designed for static tasks & No, numerical methods and adaptive correction & No, heuristic constraints only \\
\hline
\textbf{Use in RL} & Suitable for generating high-quality RL training data & Moderate, used as initial dataset & Low, designed for fixed tasks & High, suitable for adaptive RL & Low, limited RL support \\
\hline
\textbf{Computational Resource Intensity} & High, due to IPOPT and symbolic optimization, but justified for complex tasks & Moderate, optimized for ROS, suitable for research & Low, simple maintenance for static tasks & High, especially in real-time tasks & Low, focused on simple calculations \\
\hline
\end{tabular}
\end{minipage}
\end{table*}
\setlength{\textfloatsep}{0.25cm}

\subsection{Precomputed motion libraries}
There are no existing works in the literature dedicated to creating motion libraries for manipulators; however, this idea has shown promising results for bipedal robots. For instance, in \cite{Fu2024, Reher2020, Reher2021}, a library of gaits with various longitudinal and lateral speeds was created for the Cassie robot and full-body humanoid robot \cite{c5} using the C-FROST framework \cite{Hereid2021}. If a specific solution was absent in the library, it was obtained using trilinear interpolation. FROST and its extension, C-FROST, are frameworks that solve control problems for hybrid systems (i.e., systems with both continuous and discrete subsystems) using the hybrid zero dynamics method \cite{Grizzle2001}. To accelerate optimization, gradients of kinodynamic and other constraints are computed symbolically.

\subsection{Collision Detection}
The most widely used collision detection algorithms based on Gilbert-Johnson-Keerthi (GJK)\cite{Gilbert1988, Cameron1997} and Minkowski Portal Refinement (MPR)\cite{Michael2013, Wang2020} methods, which use support mapping to determine collisions between convex objects.  Enhancements to MPR and zero-crossing handling functions are proposed in \cite{Neumayr2017}. These methods are implemented in the Flexible Collision Library (FCL) \cite{Pan2012} and physics engines such as Bullet \cite{Coumans2015}, MuJoCo \cite{Todorov2012} etc. However, they are not differentiable due to logical control flow.

Differentiable Collision (DCOL) method is implemented in the Julia DifferentiableCollisions library  \cite{Tracy2024}. It determines collisions between convex primitives, including polyhedra, capsules, cylinders, cones, ellipsoids, and padded convex polygons. These shapes can approximate the geometry of the robot and any surrounding objects. DifferentiableCollisions computes the scaling factor of convex bodies at which a collision occurs. If the scaling factor is greater than one, no collision is present. The library also computes the derivative of this factor with respect to the geometric parameters of convex bodies, such as position, orientation, and size, making it suitable for gradient-based optimization methods. Thus, it has greater computational efficiency then alternative methods. 

\section{PROBLEM STATEMENT}

We consider manipulator movement with collision avoidance as a nonlinear programming problem (NLP). To describe the dynamics of the whole robot the next second-order Euler-Lagrange equation is used
\begin{equation} \label{GrindEQ__1_}
M\left(q\right)\ddot{q}+C\left(q,\ \dot{q}\right)\dot{q}+G\left(q\right)=Bu,
\end{equation}
where $q \in \mathbb{R}^n$ are the generalized coordinates, $M(q)$ is the mass matrix, $C\left(q,\ \dot{q}\right)\dot{q}$ are vectors containing the centrifugal and Coriolis forces, $G(q)$ are the gravitational forces, and $B\in \mathbb{R}^{n\times m}$ is the actuation matrix, $u\in U\subset \mathbb{R}^m$ is the control (torques) actuating the system, \textit{U} is the set of admissible control inputs.
Let's rewrite (\ref{GrindEQ__1_}) as $$
\dot{x}=f(x)+g(x) u$$ 
where  $f(x)=\left[\begin{array}{c}
\dot{q} \\
-M(q)^{-1}(C(q, \dot{q}) \dot{q}+G(q))
\end{array}\right], g(x) = \left[\begin{array}{c}
0 \\
M(q)^{-1} B
\end{array}\right] u$, $x=\left[q^T \dot{q}^T\right]^T \in \mathbb{R}^{2 n}$. The cost function: 
\begin{equation} \label{NLP}
 \mathop{\min } \limits_{x_{i} , u_{i} } \sum_{i=1}^{N}  Cost(x_{i},u_{i},i), 
\end{equation}

there are the following constraints:
\begin{itemize}
 \item \textit{C1}. The closed system dynamics equation, $\dot{x}=f\left(x\right)+g(x)u(x)$,
 \item \textit{C2}. The physical feasibility condition and fixed-joints constraints (holonomic), including the torque limits, joints angles and velocities  limits, etc.,
 \item \textit{C3}. Collision constraints, 
 \item \textit{C4}. Target constraints (end-effector final point or path).
\end{itemize}

In our work we use the sum of squares of controls as a cost function, that is $Cost  = u^{T} u$. To construct the motion library, it is necessary to solve the problems \ref{NLP} for the given sets of initial and final states, corresponding to the initial and final positions of the robot manipulators.

\section{MOTION LIBRARY DESIGN}
\subsection{NLP Solver Framework}

The FROST framework was selected for solving the NLP, as it allows for the rapid formulation and solution of control problems for robotic systems. Additionally, it enables solving control problems for hybrid systems (e.g., walking robots) in the future. FROST is implemented in MATLAB and uses the direct collocation method for numerically solving the nonlinear dynamics \textit{C1}. It supports multiple solvers. In the current work, IPOPT was used \cite{c6}. To run IPOPT from FROST, all NLP constraints must be translated into C language files and compiled using the mex compiler embedded in MATLAB. The resulting solution is not globally optimal and includes elements of stochastic search.

\subsection{Constraints}

The constraints related to the robot's dynamics and kinematics (i.e., constraints \textit{C1} and \textit{C2}) are automatically generated in FROST from the URDF file in symbolic form. The constraints \textit{C3} and \textit{C4} are user-defined based on the specific requirements of the task. FROST does not have a convenient collision detection and avoidance mechanism that is why additional third-party packages need to be used.

\subsection{Collision Detection Tool}
To handle collisions, the DifferentiableCollisions (DCOL) library implemented in Julia was used. It also provides the first derivative of the collision parameters with respect to object position and orientation, making it suitable for integration into
gradient-based control and learning methods.

\begin{figure}
    \centering
    \includegraphics[width=.48\textwidth]{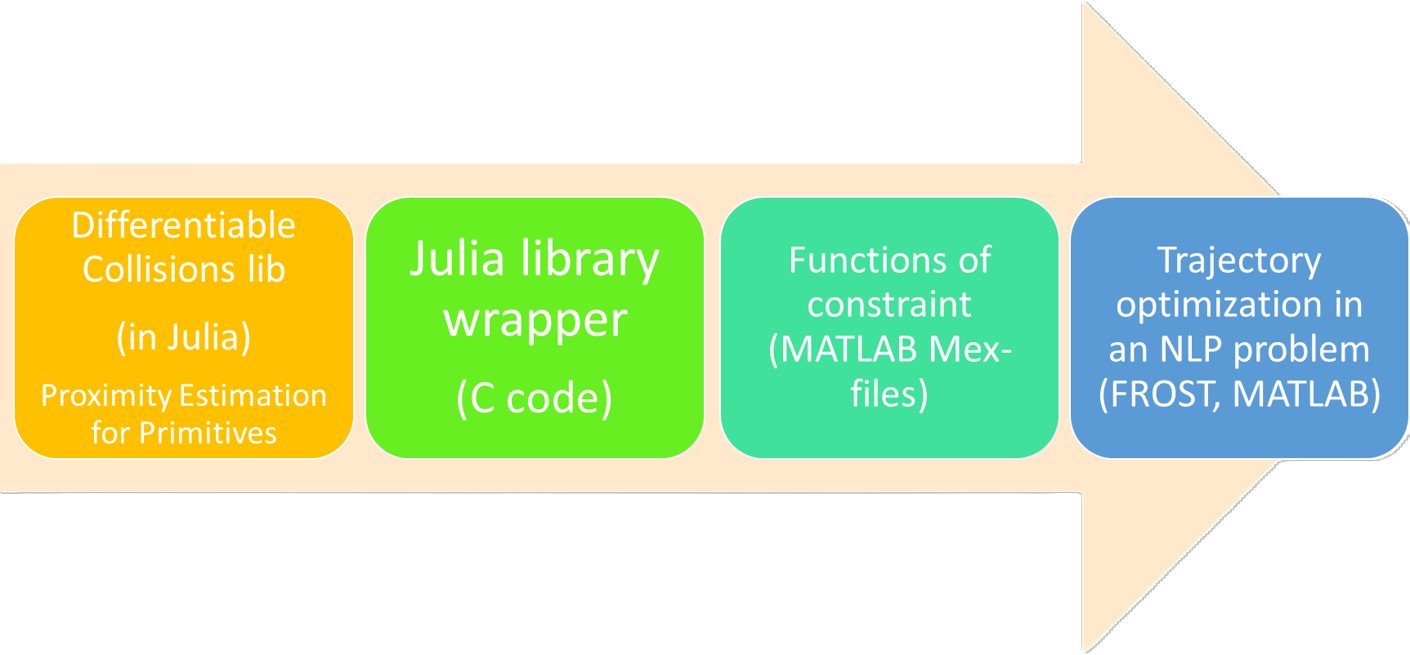}
    \caption{Adding a Collision Avoidance Constraint to the Aloha Manipulator Trajectory Optimization Problem }
    \label{fig:AddingCollisionConstraintToNLP}
\end{figure}

For convex primitive shapes, collision constraints are generated in symbolic form using the DCOL, written in Julia. For DCOL, a wrapper is generated in C, which is compiled into a MEX file, accessible for calling from FROST (MATLAB, IPOPT). The process of adding constraints to the NLP is shown in Fig.~\ref{fig:AddingCollisionConstraintToNLP}.




\subsection{Interpolation}
\begin{figure}[ht]
    \centering
    \includegraphics[width=0.48\textwidth]{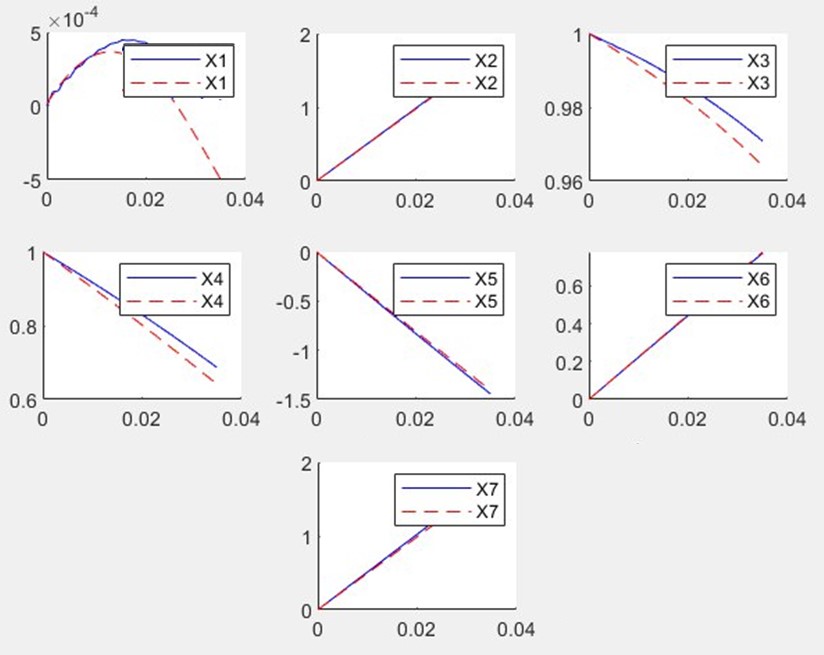}
    \caption{Interpolation errors for all joints of the Aloha manipulator when moving the manipulator from a given initial point to a final point located 40 cm away. The red dashed curve represents the interpolation results, while the solid blue line shows the exact solution obtained through optimization.}
    \label{fig:AlohaInterpolationErrors}
\end{figure}

Trilinear interpolation is employed as an alternative to solving complex optimal control problems, using 5th-order Bézier polynomials to describe the trajectories in the 8 nearest points. 
The coefficients of the polynomials for the stored trajectories are involved in the interpolation and help to determine an approximate solution to the problems not included in the library.
Trilinear interpolation is a method of multivariate interpolation on a 3-dimensional regular grid. It approximates the value of an intermediate point $(x, y, z)$ within the local axial rectangular prism linearly, using data on the lattice points. An example of interpolation errors is shown in Fig.~\ref{fig:AlohaInterpolationErrors}.

\section{Experiments}
To obtain an accurate mathematical model we have firstly identified the robot parameters (Table \ref{tab:joint_bounds_degrees}). 

\begin{table}
\centering
\caption{Joint Efforts, Velocities, and Bounds (Degrees)}
\label{tab:joint_bounds_degrees}
\begin{tabular}{llll}
\toprule
         Name & Effort (Nm) & Velocity (rad/s) & Bounds (rad) \\
\midrule
    joint1 &     9 &   3.7699 & [-2.11...3.1294] \\
    joint2 &     9 &   3.7699 & [0.044...3.65] \\
    joint3 &     9 &   3.7699 & [0.0364...3.229] \\
    joint4 &     3 &   12.5664 & [-1.37...1.36] \\
    joint5 &     3 &   12.5664 & [-1.534...1.541] \\
    joint6 &     3 &   12.5664 & [-2.1...2.087] \\
    joint7 &     3 &   12.5664 & [0.02...14.35] \\
    joint8 &     3 &   12.5664 & [0.02...14.35] \\
\bottomrule
\end{tabular}
\end{table}

\begin{figure}[h]
    \centering
    \includegraphics[width=.5\textwidth]{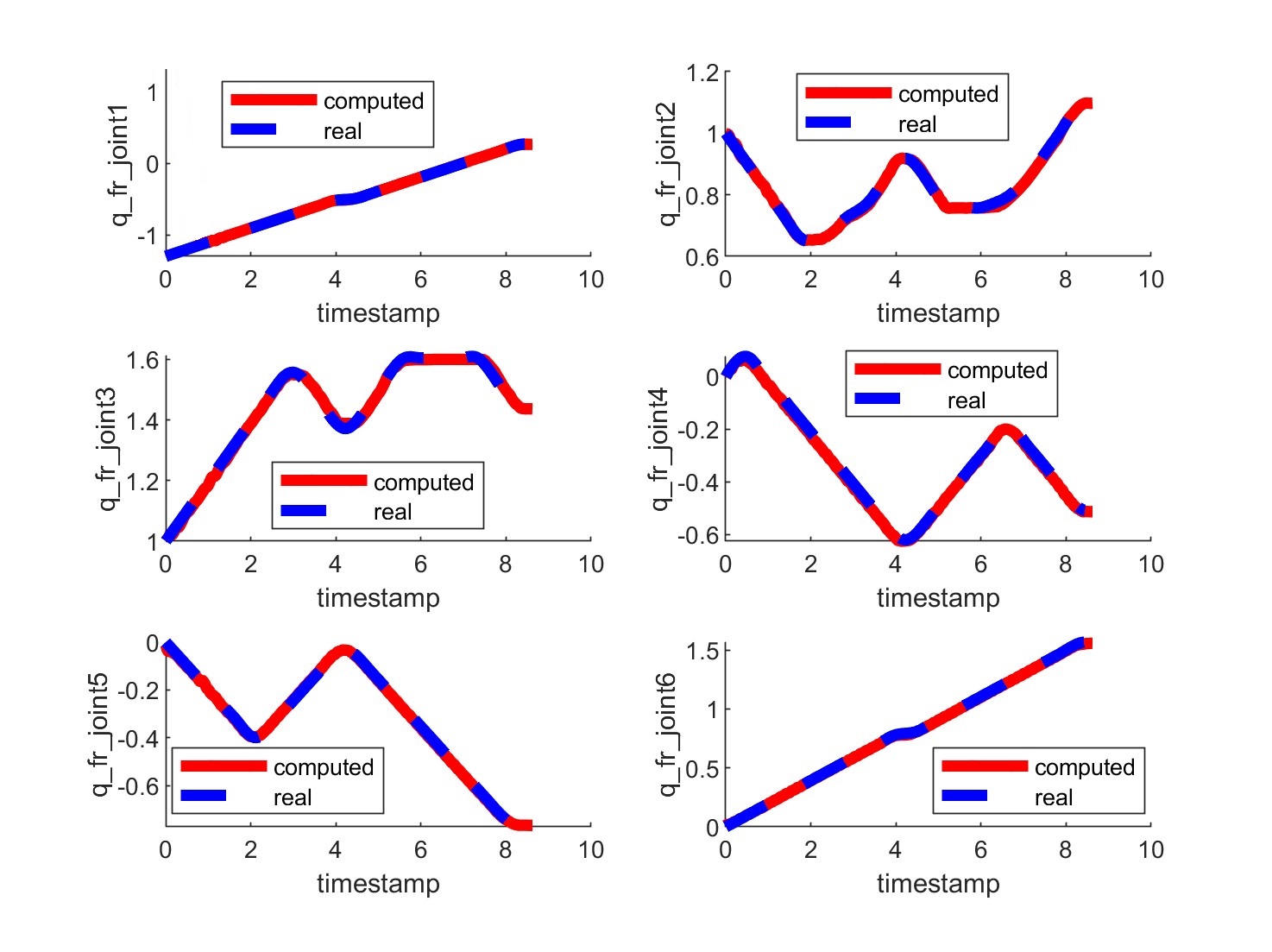}
    \caption{The comparison of real and computed joint angle trajectories}
\label{fig:AlohaTrajectoriesComparison}
\end{figure}

Secondly, we have experimentally verified that the error between the real trajectories obtained using the build-in Aloha controller and the trajectories simulated using mathematical models is quite small (Fig.\ref{fig:AlohaTrajectoriesComparison}). The average error of the joints angles is 0.0258 rad, the average error of the end-effector position in meters is 0.0170.

For the numerical experiments with TOCALib, the NLP problems (\ref{NLP}) were solved, considering the full model of Mobile Aloha with 39 DOF, i.e., $q \in \mathbb{R}^{39}$. The cost function was defined as the sum of the squared controls, and the entire time interval was divided into 31 segments (i.e. $N = 31$).

\subsection{Enhanced URDF Loading Mechanism}

The original URDF model of ALOHA was initially written as a single, monolithic file. The authors introduced a parameterization of the URDF model fields by using XACRO files and implemented a modular system \footnote{\url{https://sites.google.com/view/tocalib?usp=
sharing}}. As a result, the model became more adaptable to changes in manipulator configurations, and code duplication was eliminated, significantly enhancing its flexibility and maintainability.

\subsection{Self-collision Constraints}
Each Aloha robot manipulator consists of 8 links, each represented by a separate capsule (highlighted in transparent blue), resulting in a total of 16 capsules for the two manipulators.

The total number of 141 self-collision constrains were used, including the self-collisions of each manipulator, the collision of one manipulator with the other and, finally, the collision of manipulators with the base. A flexible mechanism was implemented for adding collision constraints between specific capsules. In the optimization process, a holonomic constraint \(\alpha(q)_i > 1, \ i = 1,2,...,141\) was applied, where \(\alpha\) is the scaling parameter for the capsules. For gradient optimization, the Jacobian for each \( \alpha(q)_i \) is computed as follows

\begin{equation} \label{Colliaion_constraint}
 \left[ \frac{\partial \alpha_i}{\partial q} \right]_{1 \times 39} = 
\left[ \frac{\partial \alpha_i}{\partial \textbf{params}} \right]_{1 \times 14} 
\left[ \frac{\partial \textbf{params}}{\partial q} \right]_{14 \times 39} ,
\end{equation}
where \textbf{params} is the parameter vector for two examined capsules which contains 3D position vectors and 4D quaternions
defining orientation (a vector of length 7 for each capsule).

\subsection{Collision with Static Objects}
For experiments, we considered the approximation of collision objects using spheres and polytopes. Each sphere is described by a radius, a position and an orientation vector. A polytope is described by a set of halfspace constraints  $Aw \le b$, where $w \in \mathbb{R}^3$, $A \in \mathbb{R}^{m \times 3}$ and $b \in \mathbb{R}^m$ represent the $m$ halfspace comprising the polytope (in our experiments we used four parallelepipeds described by $m=6$ halfspaces). A polytope also has position and orientation parameters. Thus, one additional constraint was added for collision of robot links with a sphere object $\alpha(q)_i > 1, \ i = 1,2,...,16$ (16 pairs, 8 for each manipulator) and another for collision with a shelf constructed from 4 polytopes (64 pairs for collision check).

\subsection{Tasks}

\begin{figure*}[ht]
    \centering
     \subfigure[Initial position]{\includegraphics[width=0.27\textwidth]{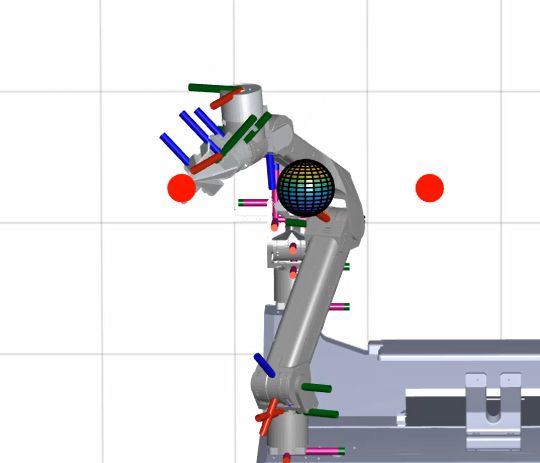}\label{fig:sub3}} 
    \subfigure[Intermediate position]{\includegraphics[width=0.2\textwidth]{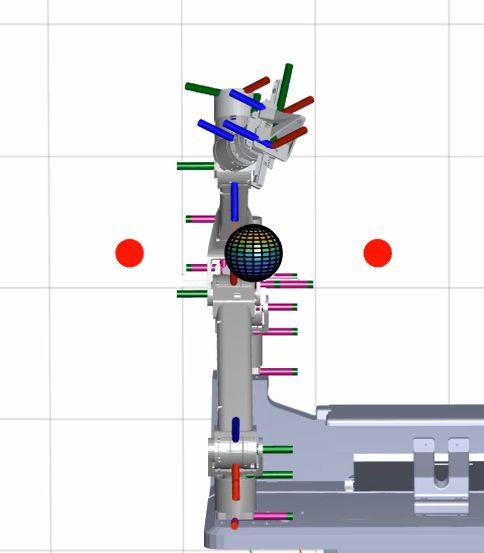}\label{fig:sub4}}
    \subfigure[Final position]{\includegraphics[width=0.26\textwidth]{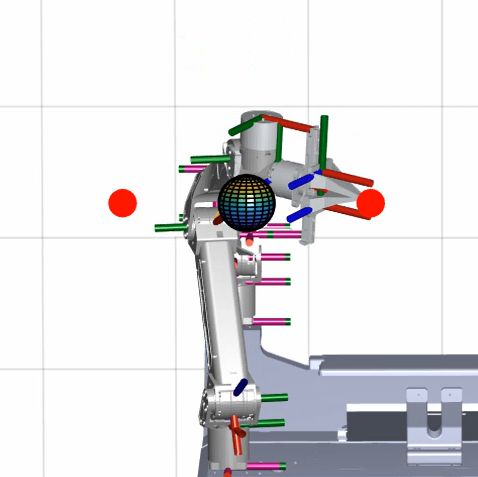}\label{fig:sub6}}
    \caption{Collision-free trajectories of a manipulator}
    \label{fig:AlohaScenarioSphere}
\end{figure*}

\begin{figure*}[ht]
    \centering
     \subfigure[Initial position]{\includegraphics[width=0.2\textwidth]{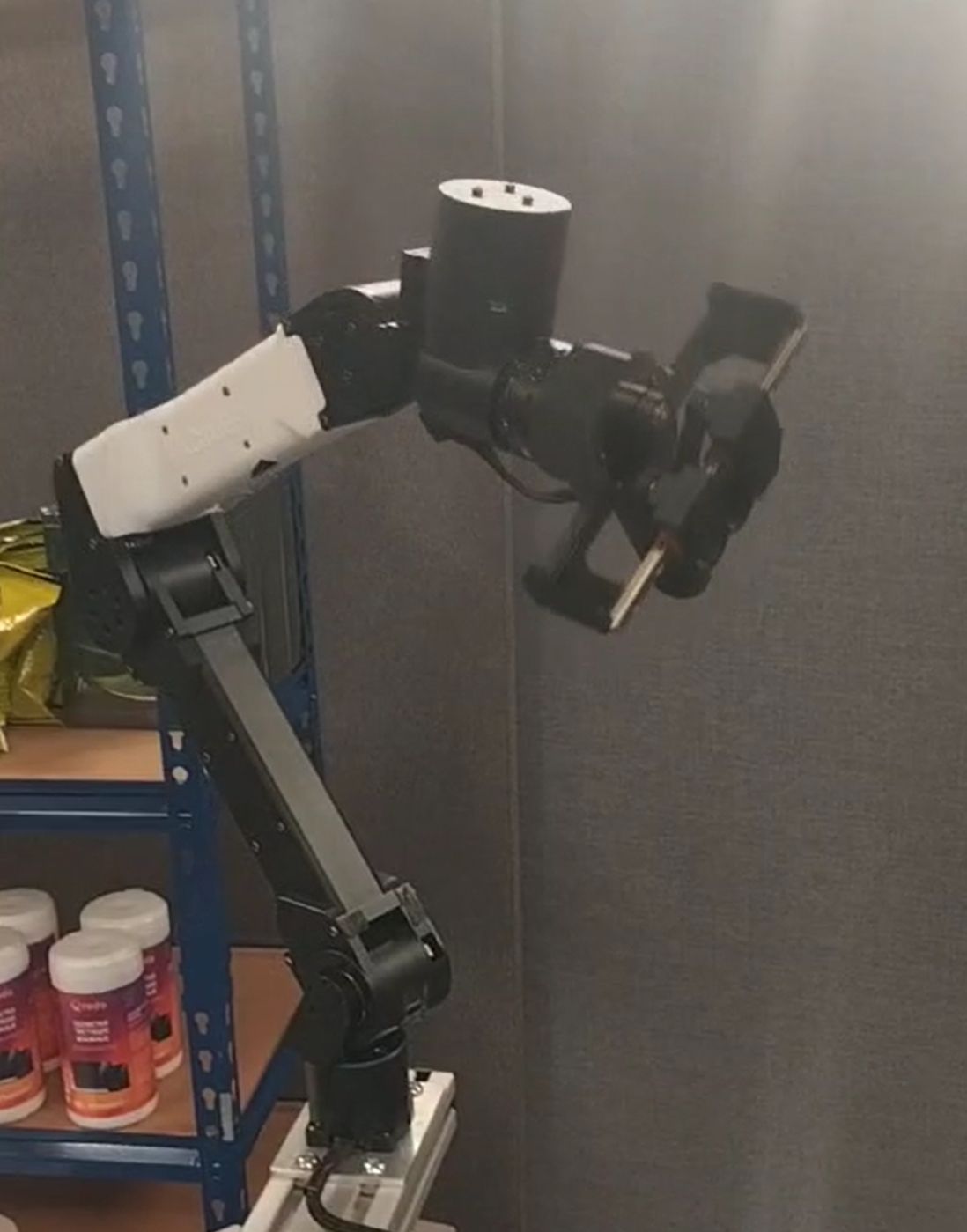}\label{fig:sub1}}
     \subfigure[Intermediate position]{\includegraphics[width=0.2\textwidth]{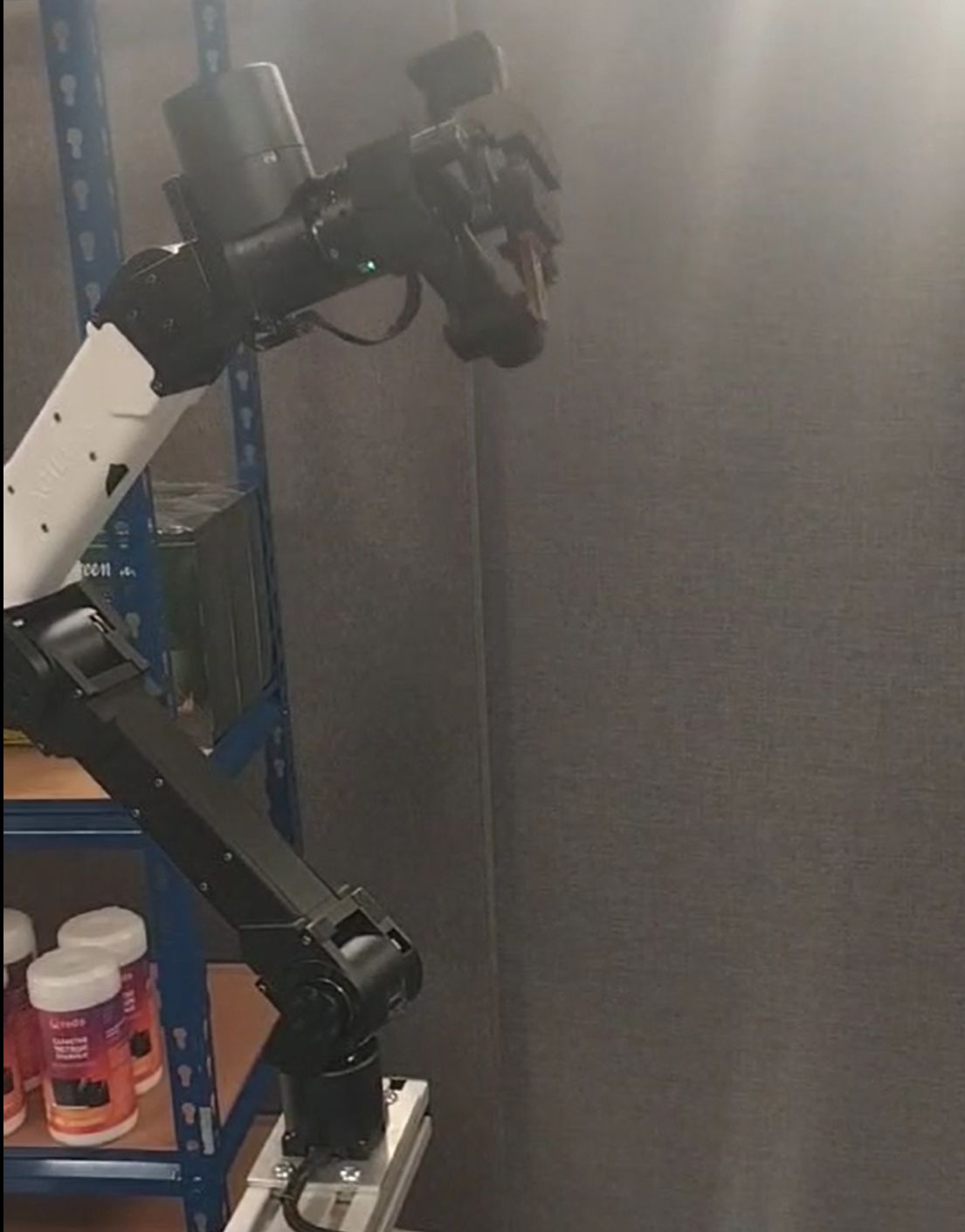}\label{fig:sub2}}
     \subfigure[Final position]{\includegraphics[width=0.2\textwidth]{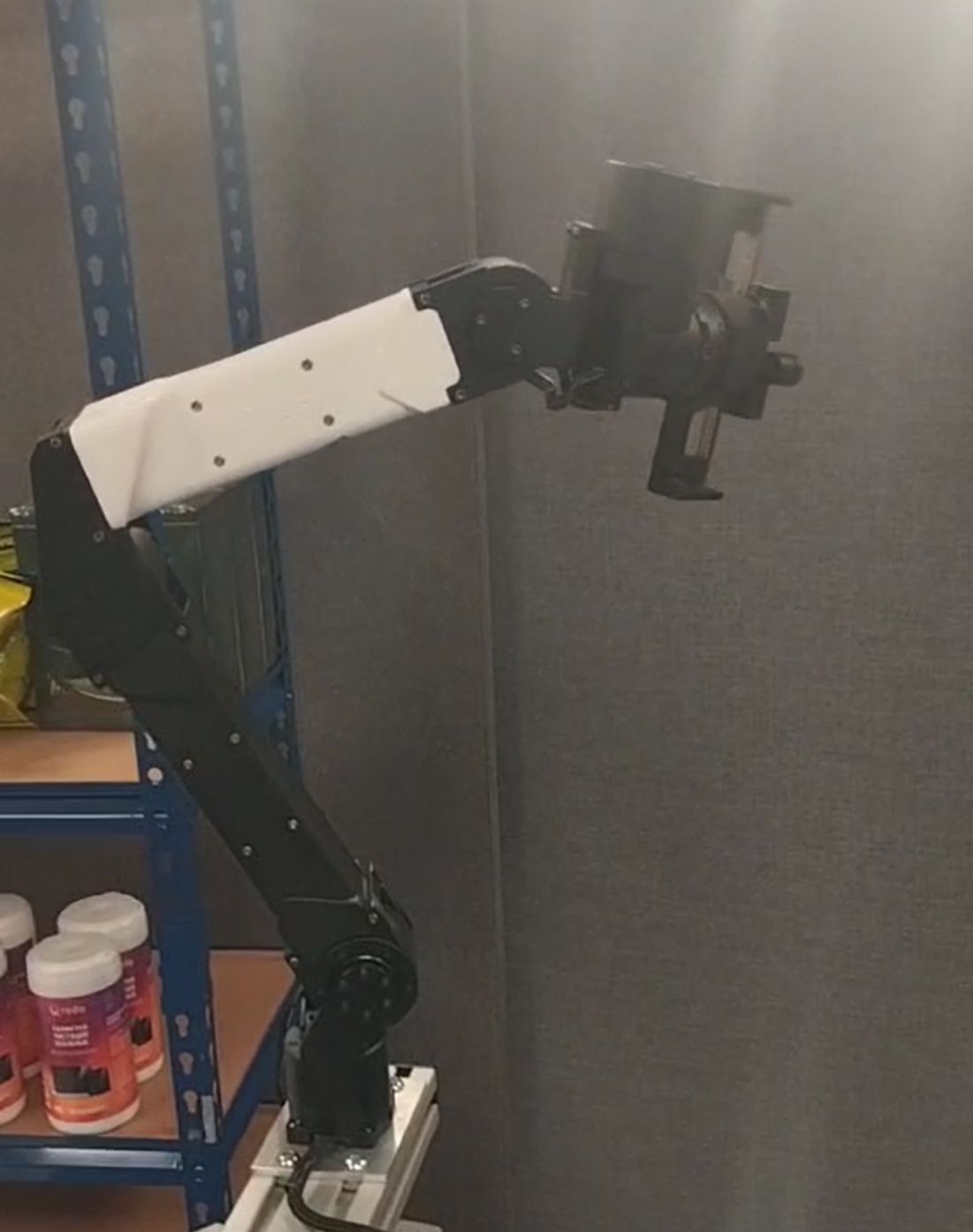}\label{fig:sub3}}
    \caption{The experiment on a real Aloha (with a virtual sphere)}
    \label{fig:AlohaScenarioReal}
\end{figure*}

\begin{figure*}[ht]
    \centering
    \subfigure[Initial position]{\includegraphics[width=0.2\textwidth]{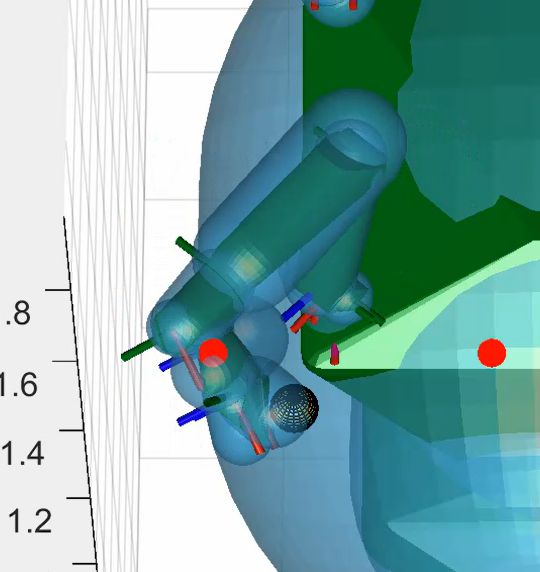}\label{fig:sub2}}
    \subfigure[Intermediate position]{\includegraphics[width=0.2\textwidth]{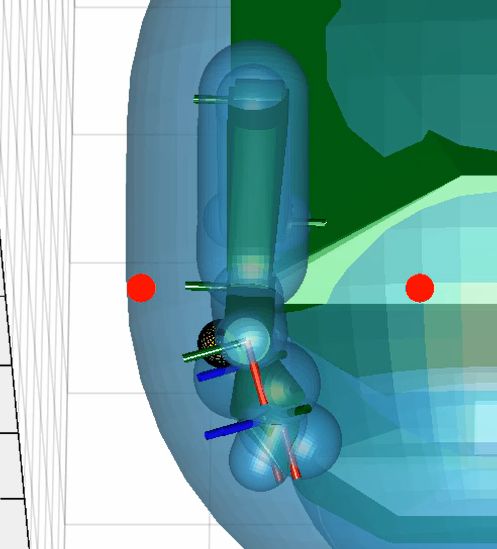}\label{fig:sub3}}
    \subfigure[Final position]{\includegraphics[width=0.2\textwidth]{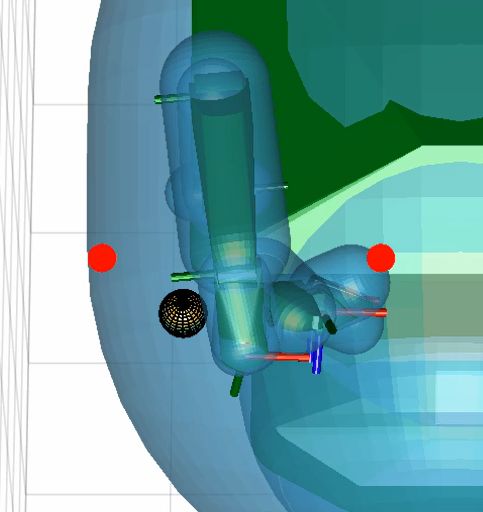}\label{fig:sub4}}
    \caption{The collision of the manipulator with the sphere without collision constraints}
    \label{fig:AlohaScenariosWithCollision}
\end{figure*}

\begin{figure*}
    \centering
    \subfigure[Initial position]{\includegraphics[width=0.2\textwidth]{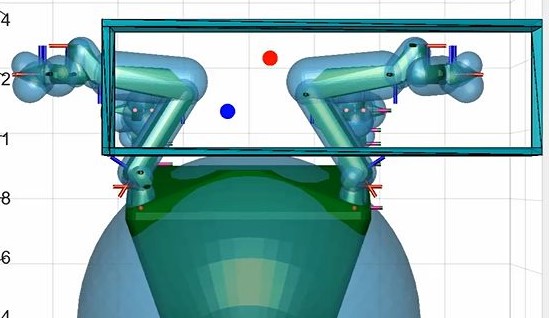}\label{fig:sub1}}
    \subfigure[Intermediate position]{\includegraphics[width=0.2\textwidth]{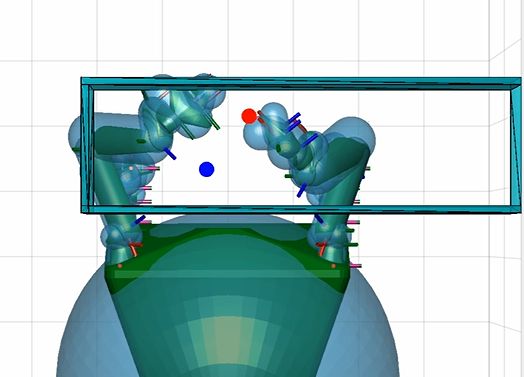}\label{fig:sub3}}
    \subfigure[Intermediate position]{\includegraphics[width=0.2\textwidth]{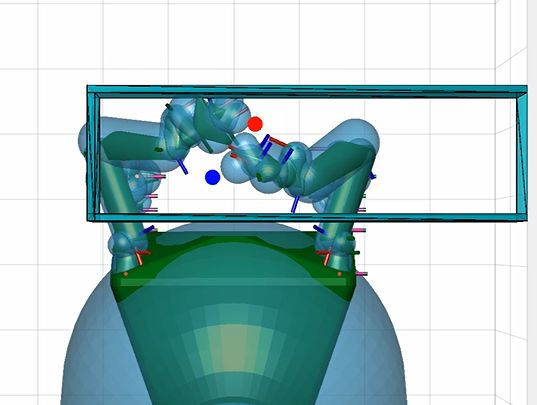}\label{fig:sub4}}
    \subfigure[Final position]{\includegraphics[width=0.2\textwidth]{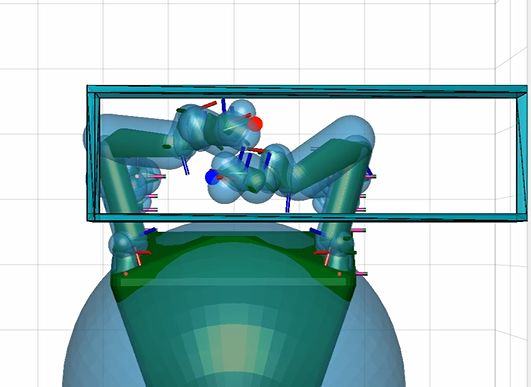}\label{fig:sub5}}
    \caption{Self collision avoidance in a presence of a shelf}
    \label{fig:SelfCollisionAvoidanceShelf}
\end{figure*}

\begin{figure*}
    \centering
    \subfigure[Initial position]{\includegraphics[width=0.2\textwidth]{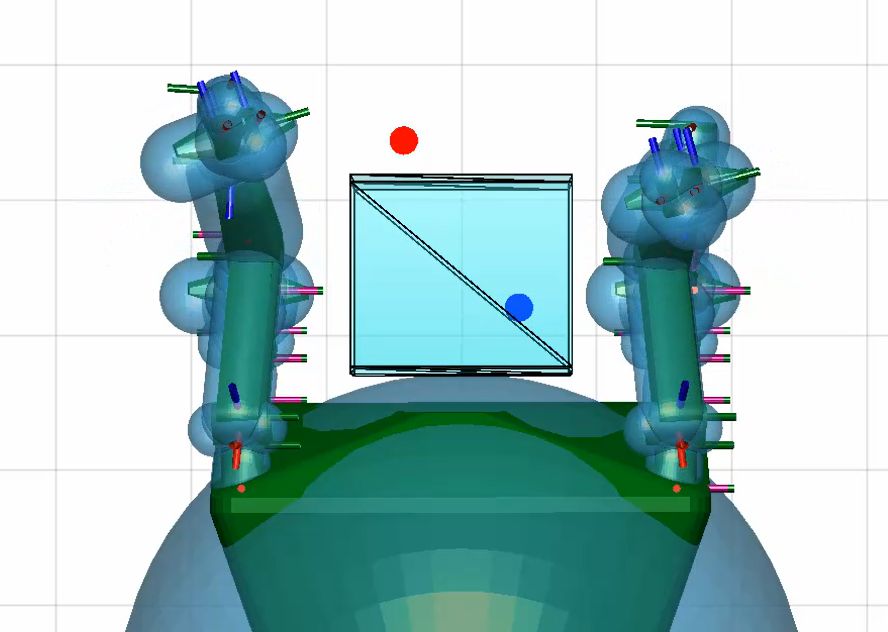}\label{fig:sub3}}
    \subfigure[Intermediate position]{\includegraphics[width=0.2\textwidth]{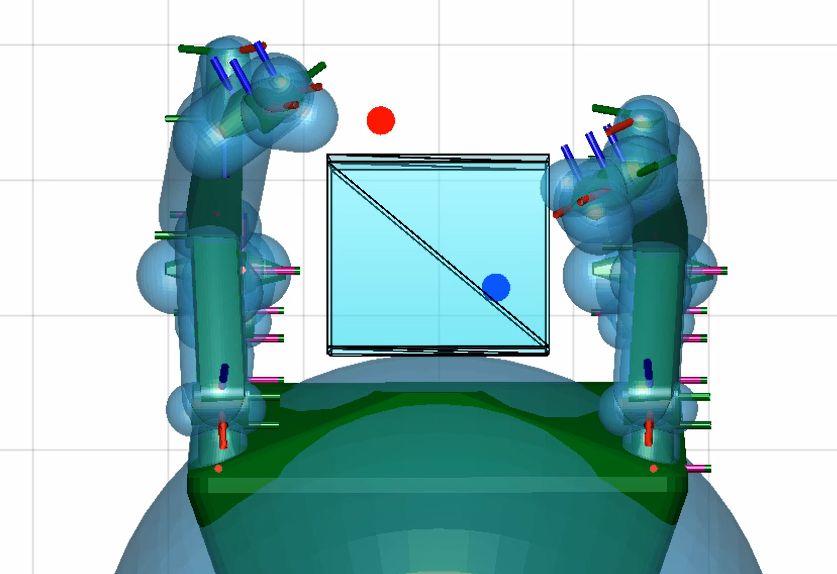}\label{fig:sub4}}
    \subfigure[Intermediate position]{\includegraphics[width=0.2\textwidth]{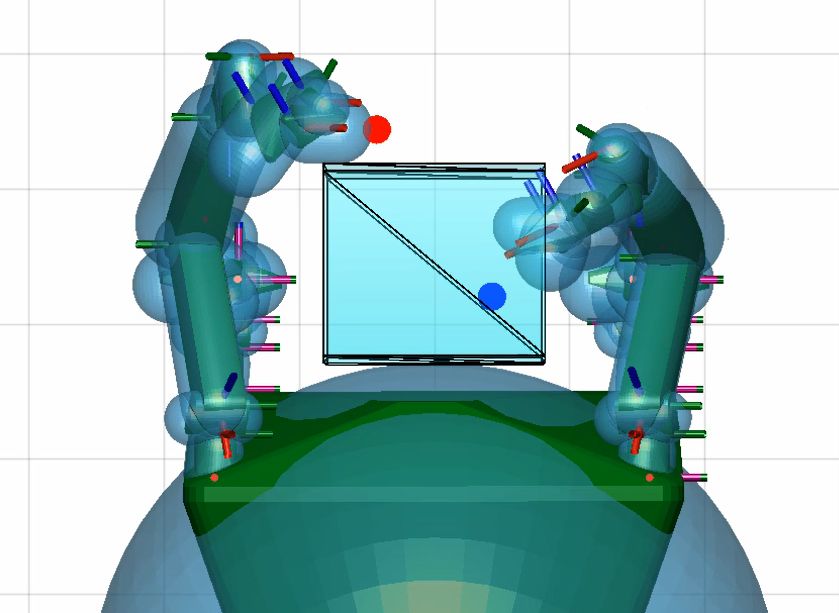}\label{fig:sub4}}
     \subfigure[Final position]{\includegraphics[width=0.2\textwidth]{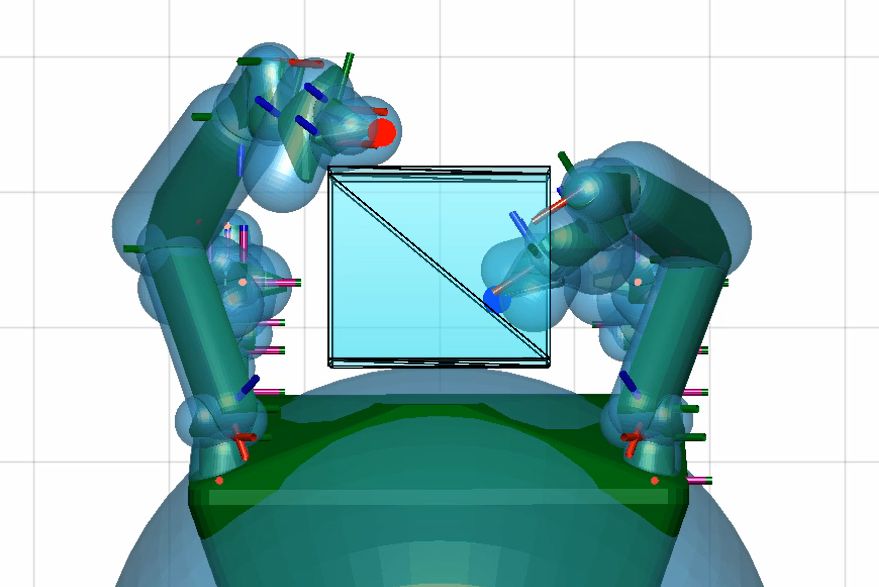}\label{fig:sub5}}
    \caption{One trajectory from a library with shelf avoidance}
    \label{fig:LibraryTrajectory}
\end{figure*}

\begin{figure*}
    \centering
    \subfigure[Initial position]{\includegraphics[width=0.2\textwidth]{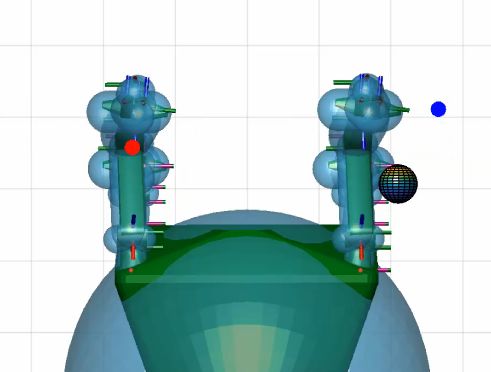}\label{fig:sub1}}
    \subfigure[Intermediate position]{\includegraphics[width=0.2\textwidth]{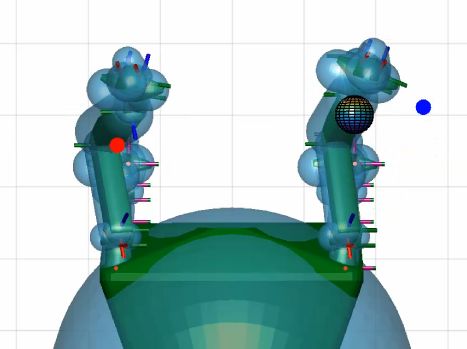}\label{fig:sub2}}
    \subfigure[Intermediate position]{\includegraphics[width=0.2\textwidth]{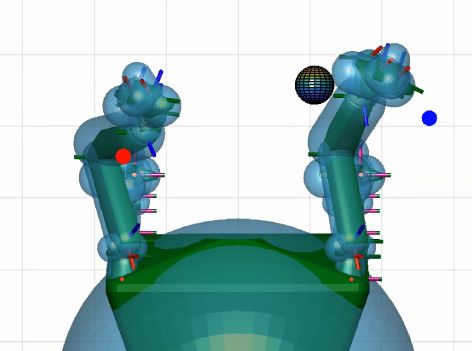}\label{fig:sub3}}
    \subfigure[Intermediate position]{\includegraphics[width=0.2\textwidth]{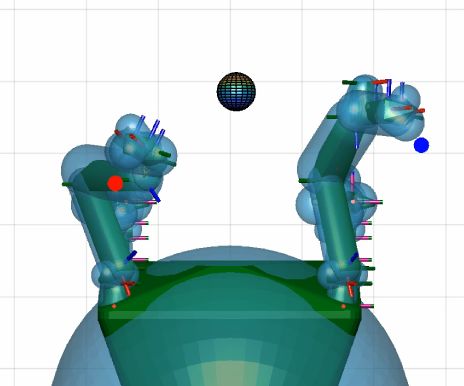}\label{fig:sub4}}
     \subfigure[Intermediate position]
     {\includegraphics[width=0.2\textwidth]{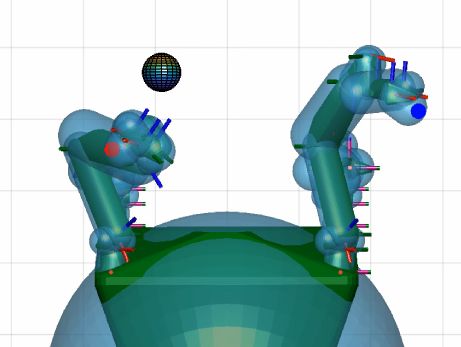}\label{fig:sub5}}
    \subfigure[Intermediate position]{\includegraphics[width=0.2\textwidth]{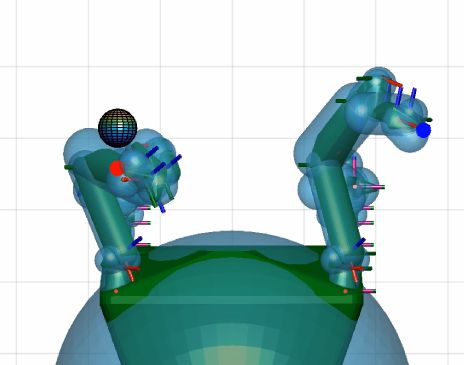}\label{fig:sub6}}
    \subfigure[Intermediate position]{\includegraphics[width=0.2\textwidth]{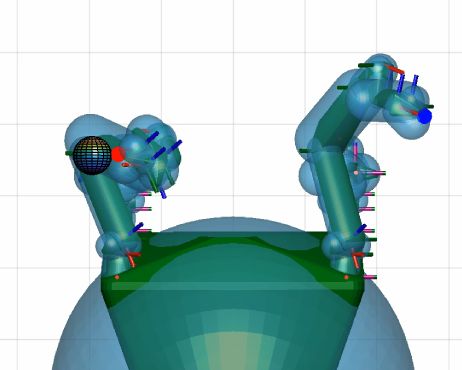}\label{fig:sub7}}
    \subfigure[Final position]{\includegraphics[width=0.2\textwidth]{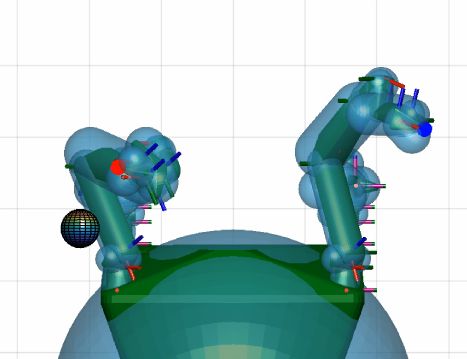}\label{fig:sub8}}
    \caption{Collision avoidance of a moving sphere}
    \label{fig:Moving_sphere}
\end{figure*}

The numerical experiments involved moving manipulators along a number of points with self-collision avoidance and collision-free moving in the presence of static objects (a sphere (Fig.~\ref{fig:AlohaScenarioSphere}-\ref{fig:AlohaScenariosWithCollision}) and a shelf (Fig.~\ref{fig:SelfCollisionAvoidanceShelf},\ref{fig:LibraryTrajectory},\ref{fig:Shelf},\ref{fig:Library})) and a moving sphere (Fig.~\ref{fig:Moving_sphere}).  The optimization tasks incorporated motion equations, self-collision constraints, and target end-effector positions. Fig. \ref{fig:AlohaScenarioSphere} corresponds to the scenario with a sphere avoidance. The goal was to follow the way-points (red dots) without collisions. The obtained trajectory was also tested on a real Aloha as shown in Fig. \ref{fig:AlohaScenarioReal}. If the simulation is carried out with the turned-off collision constrains the manipulator collides with the sphere (Fig. \ref{fig:AlohaScenariosWithCollision}). 


\begin{figure*}[h]
    \centering
    \subfigure[Initial position]{\includegraphics[width=0.2\textwidth]{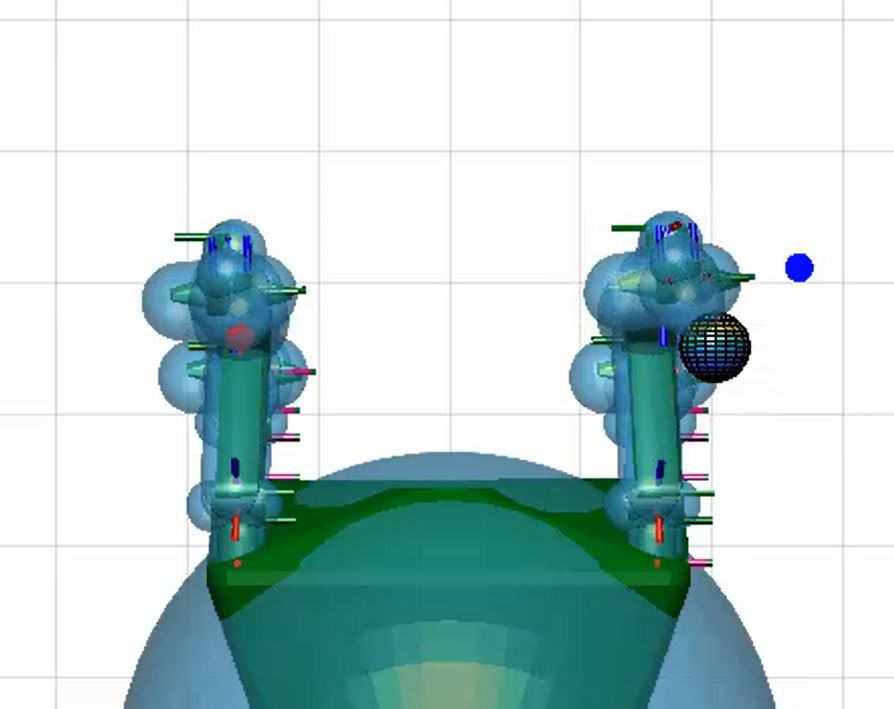}\label{fig:sub1}}
    \subfigure[Intermediate position]{\includegraphics[width=0.2\textwidth]{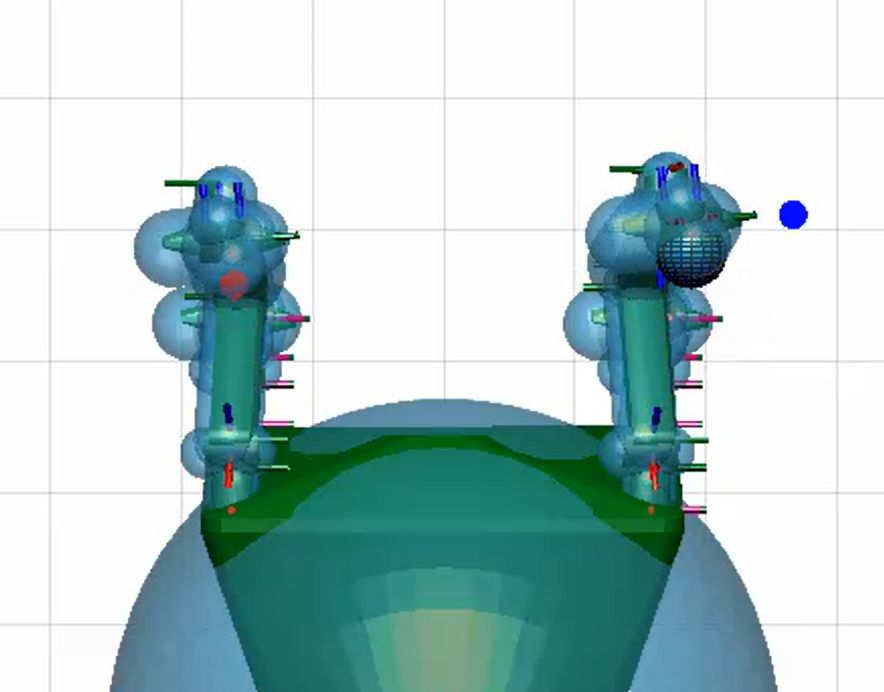}\label{fig:sub2}}
    \subfigure[Intermediate position]{\includegraphics[width=0.2\textwidth]{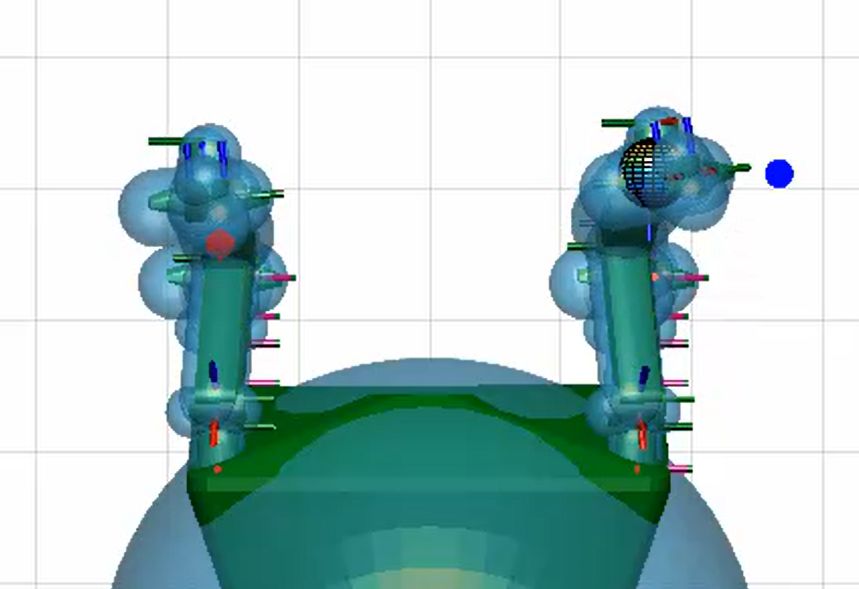}\label{fig:sub3}}
    \subfigure[Intermediate position]{\includegraphics[width=0.2\textwidth]{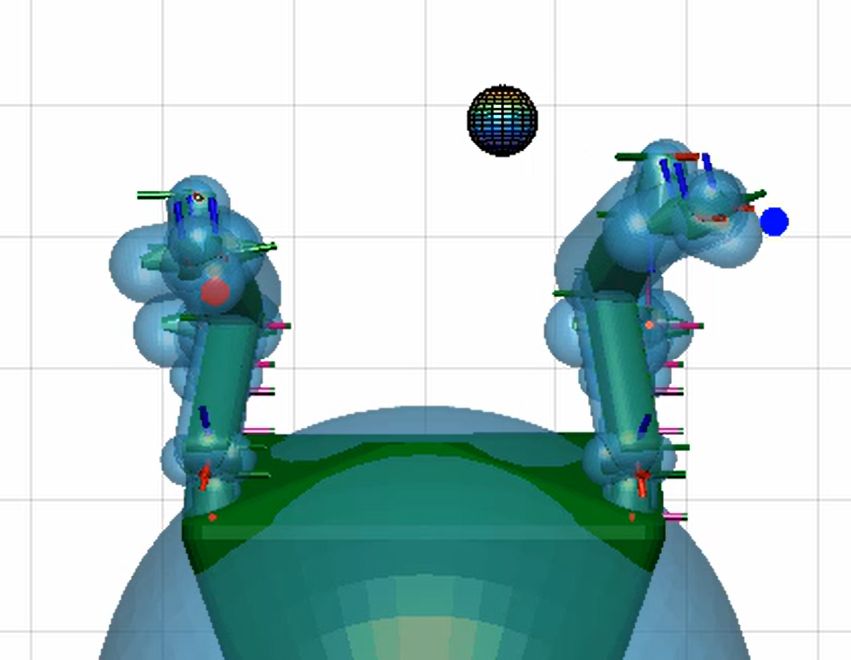}\label{fig:sub4}}
     \subfigure[Intermediate position]   {\includegraphics[width=0.2\textwidth]{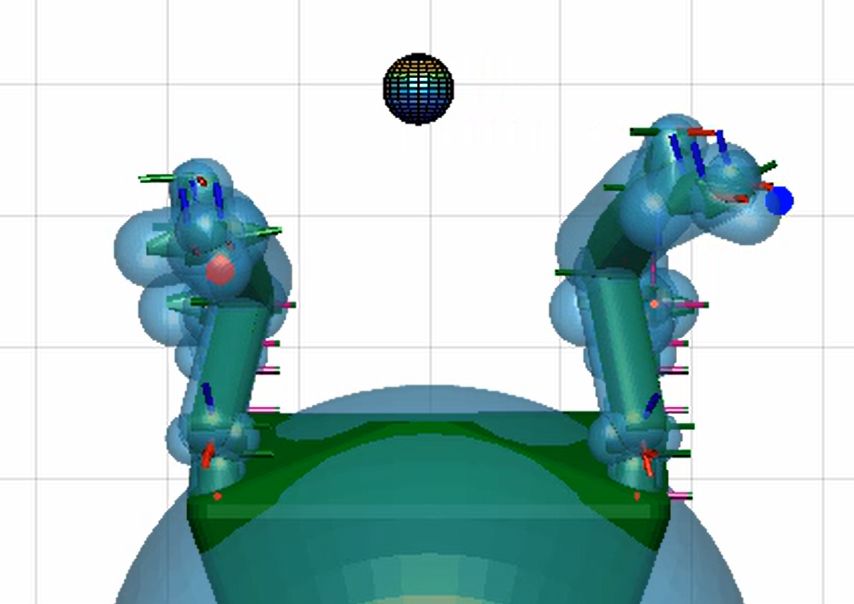}\label{fig:sub5}}
    \subfigure[Intermediate position]{\includegraphics[width=0.2\textwidth]{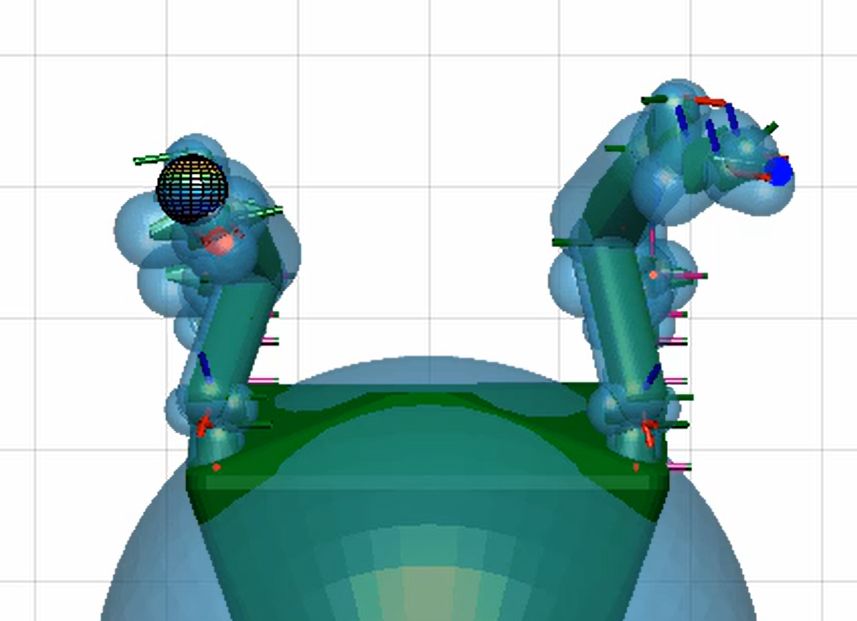}\label{fig:sub6}}
    \subfigure[Intermediate position]{\includegraphics[width=0.2\textwidth]{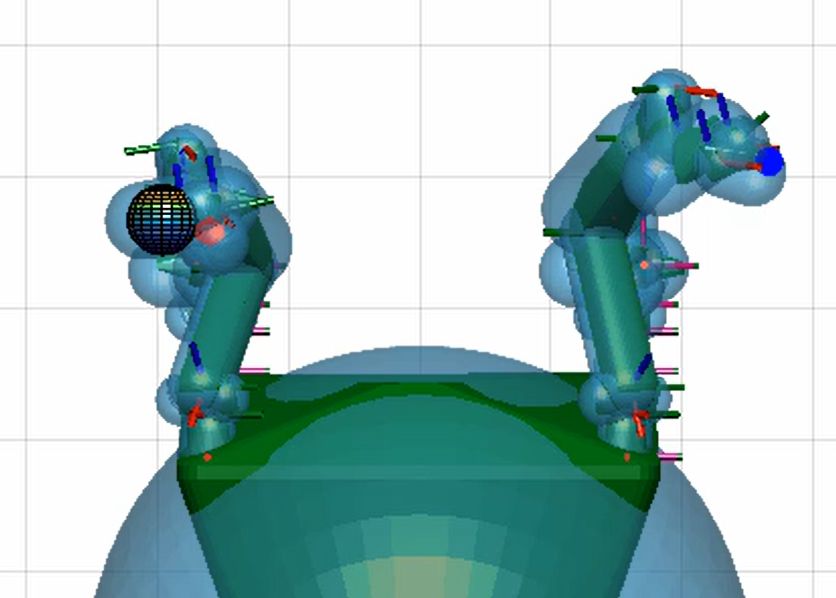}\label{fig:sub7}}
    \subfigure[Final position]{\includegraphics[width=0.2\textwidth]{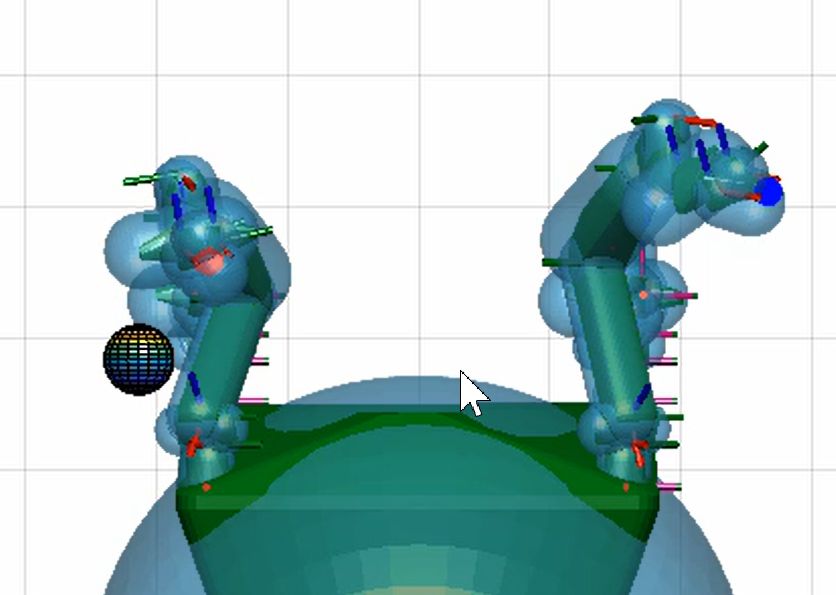}\label{fig:sub8}}
    \caption{Collision with a moving sphere without collision constraints}
    \label{fig:Moving_sphere_collision}
\end{figure*}

We have also conducted the experiments with a moving sphere (Fig.~\ref{fig:Moving_sphere}). Robot successfully avoids it when optimization is made with our collision constraints and collides with it otherwise (Fig.~\ref{fig:Moving_sphere_collision}).

The experiment with a more complicated object (Fig.~\ref{fig:Shelf}) showed that the proposed algorithm found a collision-free trajectory and the corresponding control. We compared TOCALib with CHOMP. In contrast, CHOMP algorithm provides only the way-points. To calculate the trajectories it approximates obstacles with spheres (Fig.~\ref{fig:Shelf}) and finds a smooth collision-free trajectory. Thought, the direct comparison of TOCALib and CHOMP is not possible, as CHOMP does not account for dynamics of the robot, the qualitative comparison of the obtained trajectories was made. 

Two manipulator libraries were constructed using TOCALib. The first one contains 36 goal positions within a shelf ($x \in [0.5,0.55,0.6]$, $y\in [0.55,0.1,0.15]$, $z\in [0.95,1.05,1.15]$). 

Our method succeeded (Fig.~\ref{fig:SelfCollisionAvoidanceShelf}) in 90$\%$ of cases (in 10$\%$ there was no solution because of the collision in the final state), while CHOMP managed to find the solution for 44$\%$ of cases, not including cases where it failed to give any result and the cases where the obtained solution was infeasible (false positive) because the integrity of the robot was compromised (Fig.~\ref{fig:Chomp}). Moreover, we had to turn off the self-collision option in CHOMP, otherwise it provided no solution.
 \begin{figure}[t]
    \centering
    \includegraphics[width=.42\textwidth]{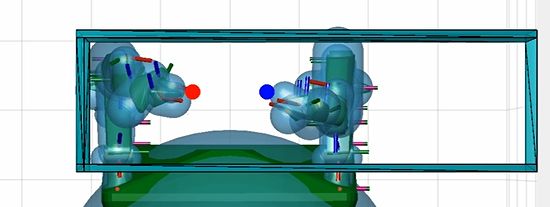}
    \caption{Collision avoidance for a shelf object}
\label{fig:Shelf}
\end{figure}
 \begin{figure}[h]
    \centering
    \includegraphics[width=.4\textwidth]{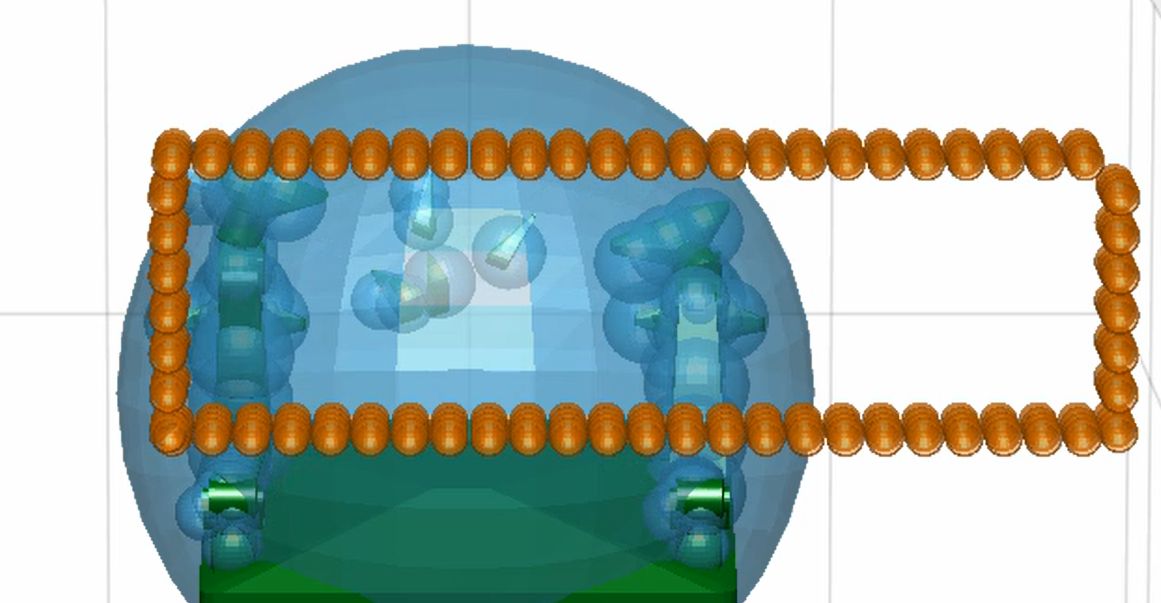}
    \caption{Approximation of a shelf in CHOMP (failed to provide a feasible trajectory: the connection of the arms is broken)}
\label{fig:Chomp}
\end{figure}

 \begin{figure}[h]
    \centering
    \includegraphics[width=.4\textwidth]{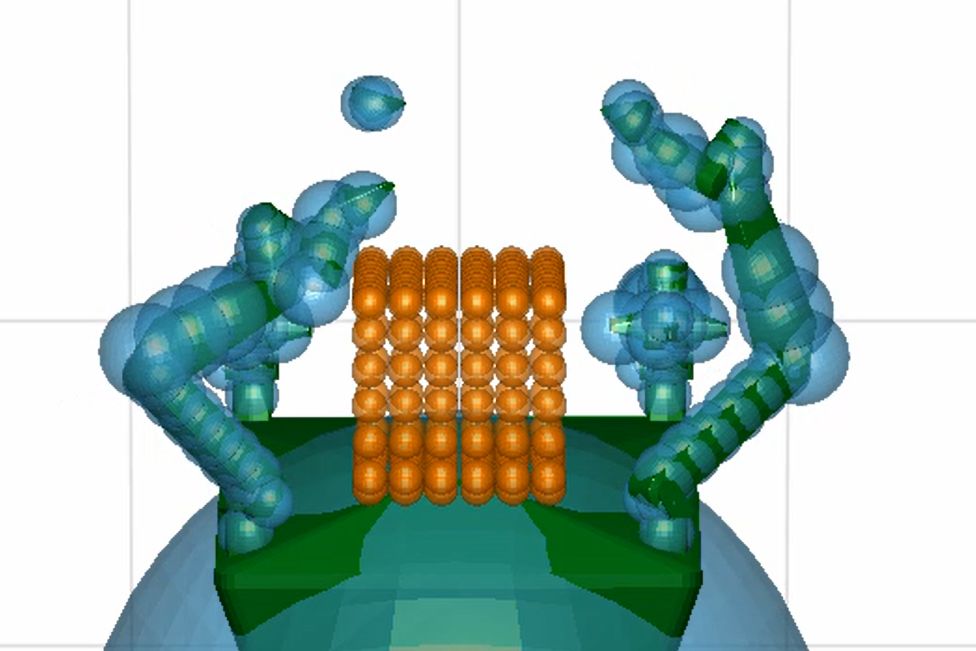}
    \caption{The trajectory obtained by CHOMP with shelf collision avoidance (one connection of the arm is broken)}
\label{fig:Chomp_shelf}
\end{figure}

 \begin{figure}[h]
    \centering  \includegraphics[width=.4\textwidth]{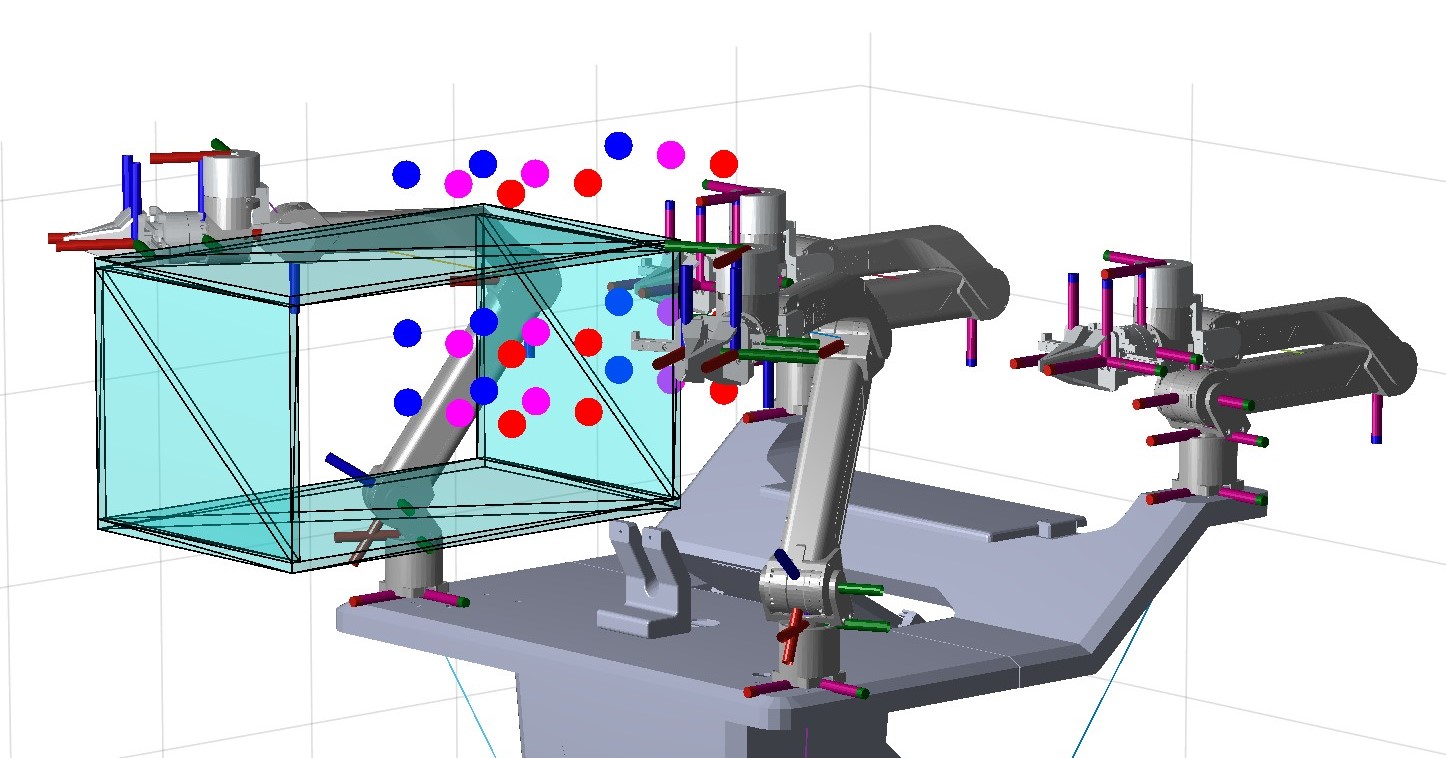}
    \caption{The distribution of goal points in a library (red for the left arm, blue - for the right arms, and magenda - for both arms)}
\label{fig:Library}
\end{figure}

The second library was constructed for an environment with a shelf placed between manipulators and 27 target positions (Fig.~\ref{fig:Library}): some of them must be reached only by one manipulator, and others must be reached by both manipulators consistently. Thus, the library includes 325 combinations of final positions for both manipulators (cases where both manipulators have the same goal positions were not considered). From 325 trajectories calculated using TOCALib (Fig.~\ref{fig:LibraryTrajectory}) and CHOMP 159 were infeasible (a task for which no solution was obtained using TOCALib or CHOMP  was considered unacceptable). The table \ref{tab:results} shows the result only for feasible cases. The CHOMP operation time was limited to 600 seconds, which corresponds to the maximum operating time limit of our method (see the section Simulation and runtime parameters). The algorithm for solving the problem using CHOMP is given in the APPENDIX.

As can be seen from the table, our approach solves the problem in a significantly larger number of cases, while spending approximately the same time as CHOMP. Moreover, in 22 \% cases the video of CHOMP trajectories shows unexpected behavior (Fig.~\ref{fig:Chomp_shelf}), thus it real  success rate is likely even lower (we did not exclude such solutions from the list of successful ones for CHOMP).

\begin{table}[h]
\caption{Results of the experiments with library}
\label{tab:results}
\centering
\begin{tabular}{|c|c|c|c|c|c|}
\hline
\textbf{Method} & \textbf{Success} & \textbf{Success rate} & \textbf{Average time, sec} \\
\hline
TOCALib  & 129 of 166 & 78\% & 483,55
\\
\hline
CHOMP & 85 of 166 & 51\% & 465,21
\\
\hline
\end{tabular}

\end{table}

\subsection{Simulation and runtime parameters}
During the simulation the end-effector target position error is set to $10^{-3}$ and the permissible deviation from the dynamic equation in \eqref{GrindEQ__1_} is $10^{-4}$. On average, trajectory computation takes approximately 2 minutes. The maximum computation time for one trajectory is set to 10 minutes. Computations were carried out on Intel Core i7 processor with 16 GB RAM.

\section{CONCLUSIONS}

The proposed Two-Arm Optimal Control and Avoidance Library (TOCALib) framework successfully combines precomputed motion strategies with collision avoidance techniques to address the challenges of bimanual manipulation. The integration of optimisation in FROST with DifferentiableCollisions method implemented in Julia for collision detection ensures the generation of high-quality motion trajectories in complex environments that are both accurate and safe. The collision constraints are symbolically represented in the optimization problem together with kinodynamic constraints  and enable the integration of efficient collision avoidance into the trajectory planning process. TOCALib offers several advantages, such as support for a wide range of robots, fast optimization using gradient-based methods, and flexible collision-checking control. Approximate solutions can also be obtained without full optimization by using interpolation which gives adaptability. TOCALib works both for static and dynamic environments. 

In our comprehensive evaluation, we demonstrated the significant advantages of TOCALib over CHOMP for trajectory planning in bimanual manipulation tasks. While CHOMP provides only waypoints without accounting for robot dynamics and approximates obstacles with spheres, TOCALib offers a complete solution with full consideration of kinodynamic constraints. Our experimental comparison using a library containing 166 feasible goal positions showed that TOCALib successfully solved 129 cases (78\% success rate), while CHOMP managed only 85 cases (51\% success rate). 

However, TOCALib has limitations. The computational time and limited number of allowed approximation primitives restrict TOCALib applicability in real-time scenarios that require rapid and accurate adaptation to environmental changes. Nevertheless, our approach provides the interpolation tool, which helps to overcome this problem by utilizing precomputed optimal feasible trajectories with close goal positions. 
Moreover, our method provides a powerful solution for creating high-quality datasets for reinforcement learning and lays the foundation for generating diverse scenarios necessary for training RL agents. The experiments demonstrated the effectiveness of TOCALib for bimanual manipulators. 

The directions of future research include the application of the method for other types of robots, implementation of parallel calculations and exploration of alternative interpolation methods.

\section{APPENDIX}
In this appendix, we present the algorithm \ref{alg:CHOMP_algorithm} used for trajectory planning with CHOMP. The \textit{CHOMPTrajectoryPlanning} procedure contains 3 main stages: 
\begin{itemize}
    \item finding a feasible trajectory for the left manipulator (steps 1-3), 
    \item finding a feasible trajectory for both manipulators (steps 4-7), 
    \item checking for self-collision between the two manipulators (step 8).
\end{itemize}

This decomposition allows for filtering out inherently infeasible trajectories. The procedure also includes a time limit for finding a single trajectory, as this can be a time-consuming process. A time limit of 600 seconds was chosen.

\textit{CHOMPTrajectoryPlanning} utilizes a grid search over Euler angles to identify feasible orientations for end-effectors using \textit{GetFeasibleState} procedure (algorithm \ref{alg:CHOMP_subalgorithm}). \textit{GetFeasibleState} returns a feasible final position that does not lead to collision with the shelf, using inverse kinematics.

\begin{algorithm*}
\caption{Trajectory Planning for Dual Manipulators using CHOMP} \label{alg:CHOMP_algorithm}
\begin{algorithmic}[1] 
\Procedure{CHOMPTrajectoryPlanning}{$x_0$,$p_{left}$,$p_{right}$}
\State \textbf{Input:} Initial robot position $x_0$, final positions of manipulators, $p_{left}$,$p_{right}$
\State \textbf{Output:} Waypoints (trajectory for $x$), status, execution time, video
\State start\_time $\gets$ CURRENT\_TIME()
\State max\_time $\gets 600$ \Comment{maximum execution time in seconds}
\State status $\gets 0$
\State $\Delta$ $\gets 0.4 \ldots$ \Comment{step size for Euler angles grid}
\While{(CURRENT\_TIME() - start\_time $<$ max\_time) \textbf{and} (status $= 0$)} 
            \State \textbf{Step 1:} Generate orientation (Euler angles) for the left manipulator
\For{$e\_x_1$ from $-\pi$ to $\pi$ with step $\Delta$} 
\For{$e\_y_1$ from $-\pi$ to $\pi$ with step $\Delta$}
\For{$e\_z_1$ from $-\pi$ to $\pi$ with step $\Delta$}

\State $a_1 \gets$ $[e\_x_1, e\_y_1, e\_z_1]$

\State \textbf{Step 2:} Inverse kinematics and collision check for left manipulator
            \State [status\_left, $x_{f\_left}$] $\gets$ 
            \Call{GetFeasibleState}{$a_1$,$p_{left}$, "left"}

          \If{status\_left $= 0$}
                \State \textbf{continue} \Comment{next iteration}
            \EndIf
            
            \State \textbf{Step 3:} Apply CHOMP algorithm for the left manipulator
            \State status\_chomp\_left $\gets$ \Call{CHOMP}{$x_0$, $x_{f\_left}$}
            
            \If{status\_chomp\_left $= 0$}
                \State \textbf{continue} \Comment{next iteration}
            \EndIf
            
            \State \textbf{Step 4:} Generate orientation for right manipulator
            \For{$e\_x_2$ from $-\pi$ to $\pi$ with step $\Delta$}
                \For{$e\_y_2$ from $-\pi$ to $\pi$ with step $\Delta$}
                    \For{$e\_z_2$ from $-\pi$ to $\pi$ with step $\Delta$}                     
                        \State $a_2 \gets$ $[e\_x_2, e\_y_2, e\_z_2]$
                        
                        \State \textbf{Step 5:} Inverse kinematics for right manipulator
                        \State [status\_right, $x_f$\_right] $\gets$ \Call{GetFeasibleState}{$a_2$,$p_{right}$, "right"}
                        
                        \If{status\_right $= 0$}
                            \State \textbf{continue} 
                        \EndIf
                        
                        \State \textbf{Step 6:} Form combined final position vector for both arms                     \State $x_f \gets$ [$x_{f\_left}$, $x_{f\_right}$]
                        
                        \State \textbf{Step 7:} Apply CHOMP algorithm for both manipulators
                        \State status\_chomp\_both $\gets$ \Call{CHOMP}{$x_0$, $x_f$}
                        
                        \If{status\_chomp\_both $= 0$}
                            \State \textbf{continue} 
                        \EndIf
                        
                        \State \textbf{Step 8:} Check entire trajectory for self-collisions
                        \State status\_self\_collision $\gets$ \Call{CheckSelfCollisions}{trajectory}
                        
                        \If{status\_self\_collision $= 1$}
                            \State status $\gets 1$
                            \State waypoints $\gets$ trajectory
                            \State \textbf{break} \Comment{exit all loops}
                        \EndIf
                    \EndFor
                \EndFor
            \EndFor
        \EndFor
    \EndFor
\EndFor
\EndWhile
\State execution\_time $\gets$ CURRENT\_TIME() - start\_time
\If{status $= 1$}
\State video $\gets$ \Call{RecordVideo}{waypoints}
\EndIf
\State \Return waypoints, status, execution\_time, video
\EndProcedure
\end{algorithmic}
\end{algorithm*}

\begin{algorithm*}
\caption{Procedure for Inverse kinematics and collision check with a polytope}\label{alg:CHOMP_subalgorithm}
\begin{algorithmic}[1]
\Statex
\Procedure{GetFeasibleState}{$a$,$p$, manipulator}
\State \textbf{Input:} $a$ orientation ($[e_x,e_y,e_z]$ euler angles), $p$ position $[x,y,z]$, manipulator name (left or right)
\State \textbf{Output:} status and $x_f$

\State \textbf{Apply inverse kinematics}
\State $x_f \gets$ \Call{InverseKinematics}{$a$, $p$, manipulator}

\If{$x_f$ does not exist}
\State \Return false, NULL
\EndIf

\State \textbf{Check collision with polytope}
\State collision $\gets$ \Call{CheckPolytopeCollision}{$x_f$} \Comment{using Julia lib and capsules approximation}

\If{collision $=$ TRUE}
\State \Return false, NULL
\Else
\State \Return true, $x_f$
\EndIf
\EndProcedure
\end{algorithmic}
\end{algorithm*}





\end{document}